# FACE SEGMENTATION: A COMPARISON BETWEEN VISIBLE AND THERMAL IMAGES

Jiří Mekyska
DT, FEEC, BUT, Brno
CZECH REPUBLIC
xmekys01@stud.feec.vutbr.cz

Virginia Espinosa-Duró
EUP Mataró
SPAIN
espinosa@eupmt.es

Marcos Faundez-Zanuy
EUP Mataró
SPAIN
faundez@eupmt.es

***Abstract*** - Face segmentation is a first step for face biometric systems. In this paper we present a face segmentation algorithm for thermographic images. This algorithm is compared with the classic Viola and Jones algorithm used for visible images. Experimental results reveal that, when segmenting a multispectral (visible and thermal) face database, the proposed algorithm is more than 10 times faster, while the accuracy of face segmentation in thermal images is higher than in case of Viola-Jones.

## I. INTRODUCTION

Face image segmentation is an important initial step for biometric face recognition [1]. While this topic has attracted a considerable attention when dealing with visible images (VIS), a large amount of topics remains unsolved for thermal images (TH). While visible images are based on the acquisition of the illumination reflected by the face, thermal images are formed thanks to the measurement of heat emission. Thus, thermal images have a nice set of interesting properties in biometric applications:

- Thermal images are not affected by illumination, shadows, etc. In fact, they can perfectly be acquired in fully darkness.
- Thermal images are more difficult to fake than visible ones, because makeup and artifacts cannot imitate the vein distribution under the skin, which is directly related to heat emission.
- Thermal images can differentiate twin brothers because heat emission is related to the veins distribution under the skin and this is different for each person, even for twin brothers.
- Thermal images contain complementary information [2] to visible images, which opens a large amount of possibilities of data fusion for enhancing pattern recognition applications.

While it is true that beard, glasses, hair, etc., can block the heat emission, in other cases the heat emission can pass through artifacts, such as plastic bags. For instance, see figure 1 and figure 2. The upper and lower images are simultaneously acquired with a visible and thermal TESTO 880-3 camera. Figure 2 shows another couple of snapshots in similar conditions. As can be observed, while it is impossible to see through the plastic bag, this is not a problem for thermal cameras. While visible camera does not show the object behind the plastic bag, the thermal image reveals that in figure 1 it is the hand without anything and that in figure 2 the hand is holding a small cup of hot coffee.

In this paper, we will propose a fast and efficient system for face image segmentation, because classic methods applied to visible images fail to segment thermal faces due to their different properties.

For comparison purposes we aply the proposed algorithm to classical visible images. In addition we also apply the classical Viola and Jones [3] algorithm for thermal face image segmentation.

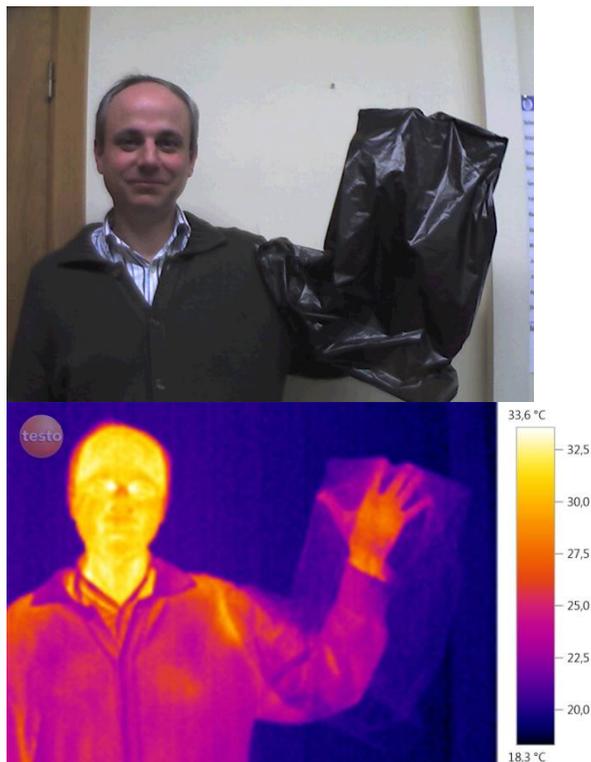

Figure 1. Example of a couple of visible and thermal images. Each pair of images has been simultaneously acquired.

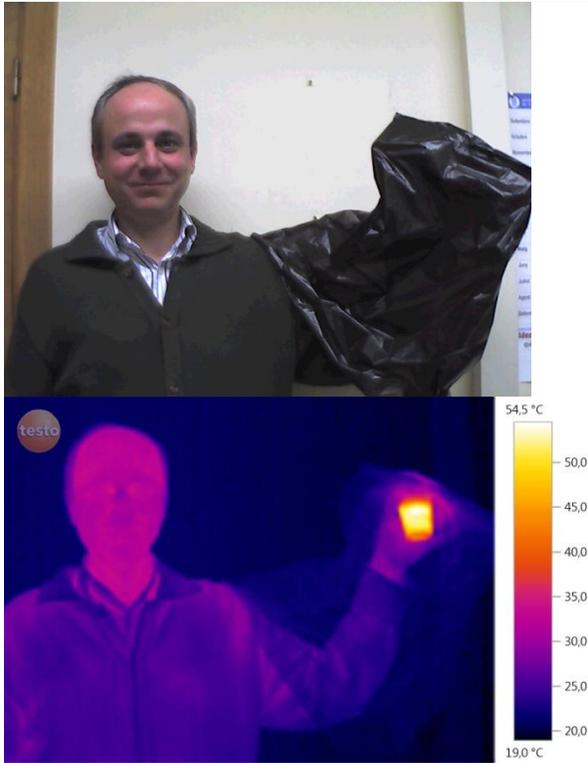

Figure 2. Example of a couple of visible and thermal images. Each pair of images has been simultaneously acquired.

## II. THERMAL FACE SEGMENTATION

In this paper we propose a segmentation method for thermal images. The goal is to detect the coordinates of the rectangle ($x_1$, $y_1$, $x_2$, $y_2$) that contain the face in a simple and efficient way. For this purpose, we propose the following algorithm:

1. The image is binarized, where the threshold $T$ is chosen by the Otsu's method [4]. If $I(x,y) < T$ then $I(x,y) = 0$, otherwise, $I(x,y) = 1$.
2. The vertical and horizontal (VH, V= Vertical, H= Horizontal) projections are calculated.
3. The first border of the vertical profile is marked as $y_1$ (see figure 2 in the middle).
4. $h/2$ is the length from the $y_1$ to the lower part of image.
5. The part of face from $y_1$ to $h/2$ is selected and then positions $x_1$ and $x_2$ are estimated as the left and right limits of this portion.
6. The lower part of the face is detected by means of $y_2 = y_1 + (x_2-x_1)*(13/9)$.

The steps of the VH projection are schematically visualized on Fig. 3, using a sample thermal image, on the top of the image. Last image shows the rectangle with the outcome of the detection process.

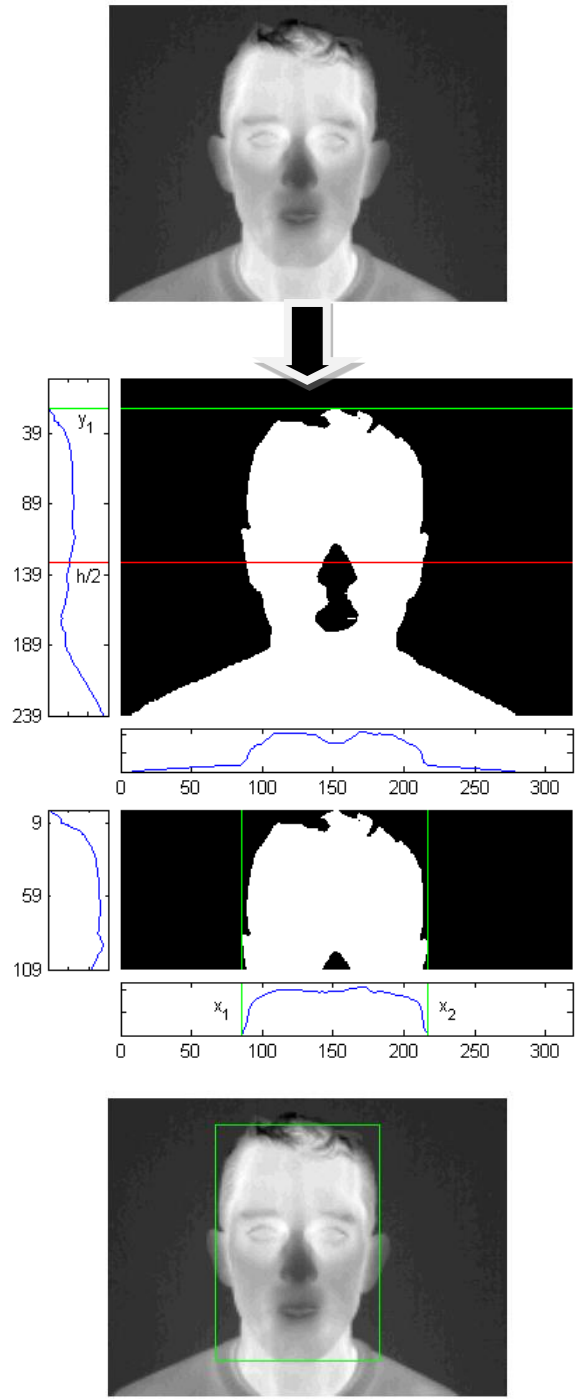

Figure 3. Algorithm steps for a sample image.

## III. EXPERIMENTAL RESULTS

In this paper, we have used a couple of VIS/TH databases:
a) Catalan database: it consists of 41 people acquired in four

different recording sessions using the thermal TESTO 880-3 thermal camera for both, visible and thermal images. It contains 2460 thermal images and 2460 visible images.

b) Italian IVITE database: it consists of 40 people acquired in a single session using the thermal TESTO 880-3 camera for thermal images and a webcam for visible images. 2400 thermal images plus 2400 visible images from this database were used.

In addition to the VH algorithm presented in section II we have also applied a Viola and Jones algorithm, which requires some training.

The experiment contains three different scenarios regarding the procedure of training Haar cascade used by Viola-Jones. In case of VH projection, there was no training which is also an advantage of this method. During the training procedure two Haar cascades were trained: one for VIS images and the other one for TH images. These scenarios are described below:

i. Scenario 1 (SC1): 900 negative (background) and 1800 positive (containing face) training images. 1800 positive images were divided into three parts containing 600 images. First part was acquired under artificial illumination (AR), second under infrared (IR) and third under natural (NA) illumination.

ii. Scenario 2 (SC2): 300 negative and 600 positive training images. These 600 positive images were acquired under artificial illumination.

iii. Scenario 3 (SC3): 1400 negative and 1800+2400 positive training images. 1800 positive images were divided into three parts containing 600 images. First part was acquired under artificial illumination, second under infrared and third under natural illumination. Next 2400 positive images were acquired under artificial illumination.

The Haar cascades were trained on frontal faces. In case of TH images, the training faces were segmented by VH projection and then manually checked or edited. In case of VIS images, the training faces were segmented by another Haar cascade trained on 7000 positive images. These faces were again manually checked or edited.

The algorithm of VH projection was written in Matlab. Viola-Jones in *.cpp and then compiled to *.mex32. The tests were run on laptop Sony VAIO VGN-NS21Z, Intel Core 2 Duo CPUP8600 2.4 GHz, 4GB RAM, MS Vista, 32-bit Matlab 2009b (version 7.9.0.529).

During the evaluation we were interested in successful detection rate (SDR) and time needed to segment the face. The SDR is defined as $100*(N_c/N_a)$ [%], where $N_c$ is number of correctly detected faces and $N_a$ is number of all faces.

Criterion of successful detection: the face must have contained at least browns, both eyes and whole lips. The face contained maximally bottom of head and in the lower parts the part of neck. Examples of badly and correctly detected faces can be seen on Figure 4 and 5.

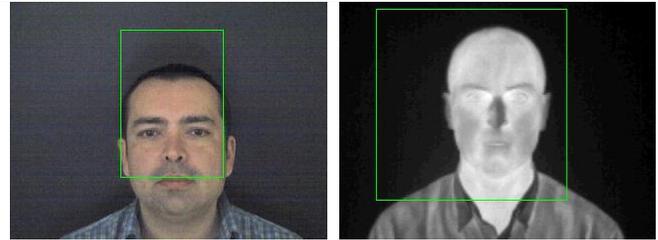

Figure 4. Examples of badly detected face in VIS and TH spectrum using Viola and Jones.

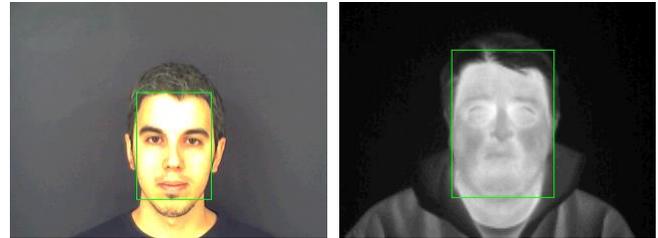

Figure 5. Examples of correctly detected faces in VIS and TH spectrums.

Figure 6 shows an example of several segmented faces. In case of first and third row, our proposed algorithm was used. In cased of second and fourth row we applied the Viola-Jones algorithm.

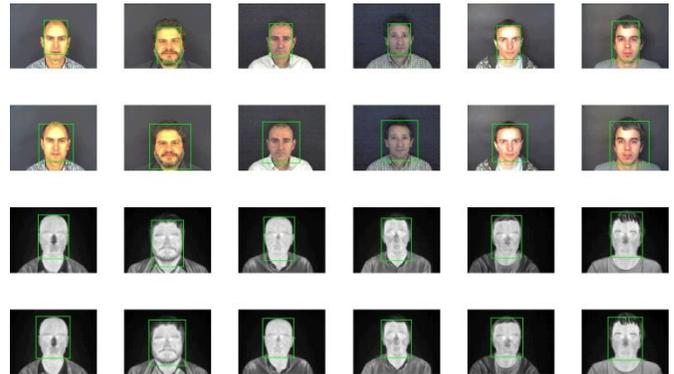

Figure 6. Some examples of photographs and the resulting face segmentation using our proposed algorithm (row 1 and 3) and Viola-Jones algorithm (row 2 and 4). Photographs in columns 1 and 2 are acquired under AR illumination, in columns 3 and 4 under IR illumination and in columns 5 and 6 under NA illumination.

Below are listed the results for all three scenarios.

|  |  | AR | IR | NA |
|---|---|---|---|---|
| VIS | VH projection | 95% | 55% | 83% |
| VIS | Viola-Jones | 51% | 48% | 46% |
| TH | VH projection | 100% | 100% | 100% |
| TH | Viola-Jones | 93% | 96% | 97% |

Table 1. (SC1) Successful detection rates.

|  |  | AR | IR | NA |
|---|---|---|---|---|
| VIS | VH projection | 95% | 55% | 83% |
|  | Viola-Jones | 40% | 42% | 40% |
| TH | VH projection | 100% | 100% | 100% |
|  | Viola-Jones | 69% | 71% | 66% |

Table 2. (SC2) Successful detection rates.

|  |  | AR | IR | NA |
|---|---|---|---|---|
| VIS | VH projection | 95% | 55% | 83% |
|  | Viola-Jones | 62% | 72% | 55% |
| TH | VH projection | 100% | 100% | 100% |
|  | Viola-Jones | 95% | 93% | 96% |

Table 3. (SC3) Successful detection rates.

|  |  | d [s] | mean(d) [ms] | var(d) [ms] |
|---|---|---|---|---|
| VIS | VH projection | 0,740 | 2,311 | 0,012 |
|  | Viola-Jones | 10,068 | 31,463 | 0,503 |
| TH | VH projection | 0,578 | 1,806 | 0,066 |
|  | Viola-Jones | 5,808 | 18,149 | 0,208 |

Table 4. (SC1) Detection time of 220 images under artificial illumination

|  |  | d [s] | mean(d) [ms] | var(d) [ms] |
|---|---|---|---|---|
| VIS | VH projection | 0,754 | 2,356 | 0,011 |
|  | Viola-Jones | 9,971 | 31,159 | 0,553 |
| TH | VH projection | 0,547 | 1,708 | 0,044 |
|  | Viola-Jones | 5,756 | 17,987 | 0,184 |

Table 5. (SC2) Detection time of 220 images under artificial illumination

|  |  | d [s] | mean(d) [ms] | var(d) [ms] |
|---|---|---|---|---|
| VIS | VH projection | 0,767 | 2,397 | 0,011 |
|  | Viola-Jones | 10,438 | 32,618 | 0,524 |
| TH | VH projection | 0,450 | 1,407 | 0,019 |
|  | Viola-Jones | 6,440 | 20,125 | 0,240 |

Table 6. (SC3) Detection time of 220 images under artificial illumination

|  |  | d [s] | mean(d) [ms] | var(d) [ms] |
|---|---|---|---|---|
| VIS | VH projection | 0,646 | 2,019 | 0,002 |
|  | Viola-Jones | 10,097 | 31,552 | 0,531 |
| TH | VH projection | 0,435 | 1,358 | 0,001 |
|  | Viola-Jones | 5,726 | 17,894 | 0,185 |

Table 7. (SC1) Detection time of 220 images under infrared illumination.

|  |  | d [s] | mean(d) [ms] | var(d) [ms] |
|---|---|---|---|---|
| VIS | VH projection | 0,674 | 2,106 | 0,003 |
|  | Viola-Jones | 9,548 | 29,838 | 0,411 |
| TH | VH projection | 0,432 | 1,350 | 0,001 |
|  | Viola-Jones | 5,884 | 18,388 | 0,227 |

Table 8. (SC2) Detection time of 220 images under infrared illumination.

|  |  | d [s] | mean(d) [ms] | var(d) [ms] |
|---|---|---|---|---|
| VIS | VH projection | 0,677 | 2,114 | 0,002 |
|  | Viola-Jones | 10,348 | 32,339 | 0,549 |
| TH | VH projection | 0,371 | 1,158 | 0,001 |
|  | Viola-Jones | 6,197 | 19,365 | 0,172 |

Table 9. (SC3) Detection time of 220 images under infrared illumination.

|  |  | d [s] | mean(d) [ms] | var(d) [ms] |
|---|---|---|---|---|
| VIS | VH projection | 0,675 | 2,108 | 0,002 |
|  | Viola-Jones | 10,178 | 31,807 | 0,528 |
| TH | VH projection | 0,430 | 1,345 | 0,001 |
|  | Viola-Jones | 5,800 | 18,125 | 0,203 |

Table 10. (SC1) Detection time of 220 images under natural illumination.

|  |  | d [s] | mean(d) [ms] | var(d) [ms] |
|---|---|---|---|---|
| VIS | VH projection | 0,692 | 2,163 | 0,003 |
|  | Viola-Jones | 9,886 | 30,895 | 0,539 |
| TH | VH projection | 0,438 | 1,368 | 0,001 |
|  | Viola-Jones | 5,702 | 17,819 | 0,165 |

Table 11. (SC2) Detection time of 220 images under natural illumination.

|  |  | d [s] | mean(d) [ms] | var(d) [ms] |
|---|---|---|---|---|
| VIS | VH projection | 0,697 | 2,180 | 0,003 |
|  | Viola-Jones | 10,404 | 32,512 | 0,524 |
| TH | VH projection | 0,376 | 1,175 | 0,001 |
|  | Viola-Jones | 6,231 | 19,472 | 0,175 |

Table 12. (SC3) Detection time of 220 images under natural illumination.

Experimental results reveal that VH projection outperform the Viola-Jones algorithm for the three kinds of image illumination in both spectrums (VIS and TH) according to computational burden (tables 4 to 12) as well as SDR (table 1 to 3). Only for IR and SC3 Viola and Jones provides higher SDR (see table 3). Nevertheless we should emphasize that classical Haar cascade training requires a higher number of face images. We could not use more due to the limitation of the used databases.

## IV. CONCLUSIONS

As was already mentioned, the advantage of VH projection is that we can leave out the training procedure. The accuracy and duration of face segmentation is thus not dependent on the amount of training data. In case of segmentation based on Viola-Jones algorithm, a large amount of training data is needed to achieve good detection rate. In this paper we trained the Haar cascades by max. 4200 images, but to obtain good results, 7000 images and more would be better. The disadvantage of this algorithm is also the detection time, which is dependent on the amount of training data. Segmentation using Haar cascade trained on many thousands images can be then more than 50 times slower than in case of VH projection.

As was mentioned in section I, thermal images are not affected by the illumination. This property is also good for the detection rate of Viola-Jones, because if the images are acquired under three different kinds of illumination, we can theoretically use for training three times less TH images than VIS images to obtain the same SDR. If we compare SDR of Viola-Jones algorithm using haar cascade trained on 600 TH images (SC2) and SDR of Viola-Jones using Haar cascade trained on 1800 VIS images (SC1) affected by illumination, the results in case of SC2 are still better.

The disadvantage of VH projection is the fact that it can detect only one face on the image while Viola-Jones algorithm can detect more than one face. Therefore VH projection is suitable for segmentation of databases containing a large number of single faces. This limitation can be in the further work cancelled by an extension to VH projection, which will firstly select the areas with potential faces and then apply VH algorithm which extracts these faces from the areas. This extension can be based on image binarization with consequent split of scene to areas according to the detected objects.

It is also interesting to comment the possibility to use the combination of VH projection and Viola-Jones algorithm. This combination was also used in this work. VH projection can be used to automatically create a large thermal training database for Viola-Jones.

## VI. ACKNOWLEDGEMENTS

This work has been supported by FEDER, MEC, TEC2009-14123-C04-04, COST OC08057, COST 2102 and KONTAKT ME 10123.

## VIII. VITA

Jiri Mekyska was born in Holesov, Czech Republic. He got the Ms.C. degree from Telecommunications in 2010 at BUT (Brno University of Technology). He deals with speech signal synthesis/analysis and thermal image processing.

Virginia Espinosa-Duró was born in Barcelona, Spain. She received the BS degree in Electrical Engineering from the Polytechnic University of Catalonia in 1992 and BSc degree in Electronic Engineering from the Barcelona University. Currently she is associate Professor in the Electronic Department at EUP Mataró in Barcelona (Spain), belongs to the signal processing research group and she leaded the Electronic and Automatic Department from 2007 to 2010. Her research interests include pattern recognition, signal processing, learning machine and neural networks. She is the vice-chair of ICCST'2011.

Marcos Faundez-Zanuy was born in Barcelona, Spain. He received the B.Sc. degree in Telecommunications in 1993 and the Ph.D. degree in 1998, both from the Polytechnic University of Catalunya. He is now associate Professor at EUP Mataró since 1994 and heads the signal processing group there. His research interests lie in the fields of biometrics and speech coding. He was the initiator and chairman of the European COST action 277 "Nonlinear speech processing". Dr Faundez-Zanuy is a Spanish Liaison for EURASIP (European Association for Signal and Image Processing). He is the dean of EUP Mataró since 2009 and the chair of ICCST'2011.